\documentclass{article}

\usepackage{arxiv}

\usepackage[utf8]{inputenc} 
\usepackage[T1]{fontenc}    

\usepackage{cite}
\usepackage{amsmath,amssymb,amsfonts}
\usepackage{algorithmic}
\usepackage{graphicx}
\usepackage{textcomp}
\usepackage{xcolor}
\usepackage{url}
\usepackage[hidelinks]{hyperref}
\usepackage{csquotes}
\usepackage{booktabs}
\usepackage{tikz}
\usepackage{cleveref}
\usepackage{isomath}
\usepackage{siunitx}
\usepackage{fontawesome}

\newcommand{\mSet}{\mathcal}

\newcommand{\mVector}[1]{\mathbf{#1}}

\newcommand{\mFunction}[1]{\mathit{#1}}

\definecolor{forestspring0}{HTML}{376819}
\definecolor{forestspring1}{HTML}{234B0E}
\definecolor{forestspring2}{HTML}{54842F}
\definecolor{forestspring3}{HTML}{80A554}
\definecolor{forestspring4}{HTML}{142F0A}
\definecolor{forestspring5}{HTML}{B6D28F}
\definecolor{forestspring6}{HTML}{E5EFDC}
\definecolor{forestspring7}{HTML}{0C1408}
\definecolor{forestspring8}{HTML}{E7EFFF}
\definecolor{forestspring9}{HTML}{F6FBEB}
\definecolor{forestspring10}{HTML}{ECEEFF}
\definecolor{forestfall0}{HTML}{BE3B16}
\definecolor{forestfall1}{HTML}{A5602F}
\definecolor{forestfall2}{HTML}{59361F}
\definecolor{forestfall3}{HTML}{4F632F}
\definecolor{forestfall4}{HTML}{414329}
\definecolor{forestfall5}{HTML}{060706}
\definecolor{forestfall6}{HTML}{322625}
\definecolor{forestfall7}{HTML}{FDCF45}
\definecolor{forestfall8}{HTML}{965520}
\definecolor{fields0}{HTML}{A8CC48}
\definecolor{fields1}{HTML}{9EC46A}
\definecolor{fields2}{HTML}{81AD31}
\definecolor{fields3}{HTML}{417217}
\definecolor{fields4}{HTML}{5F9025}
\definecolor{fields5}{HTML}{2A5716}
\definecolor{fields6}{HTML}{576828}
\definecolor{fields7}{HTML}{A2BF9F}
\definecolor{coastal0}{HTML}{EBE8E2}
\definecolor{coastal1}{HTML}{7A8A8F}
\definecolor{coastal2}{HTML}{A8B9C5}
\definecolor{coastal3}{HTML}{65584D}
\definecolor{coastal4}{HTML}{C6B7A7}
\definecolor{coastal5}{HTML}{917663}
\definecolor{coastal6}{HTML}{97A78F}
\definecolor{coastal7}{HTML}{BBD5DE}
\definecolor{coastal8}{HTML}{FFFFFF}
\definecolor{coastal9}{HTML}{FFFFFF}
\definecolor{coastal10}{HTML}{DCF4FC}
\definecolor{arctic0}{HTML}{BBBFCB}
\definecolor{arctic1}{HTML}{BAB9B7}
\definecolor{arctic2}{HTML}{949CA6}
\definecolor{arctic3}{HTML}{F0F5FF}
\definecolor{arctic4}{HTML}{F6FBFF}
\definecolor{arctic5}{HTML}{EBEEEF}
\definecolor{arctic6}{HTML}{59626F}
\definecolor{arctic7}{HTML}{F6F6F6}
\definecolor{arctic8}{HTML}{FFFFFF}
\definecolor{arctic9}{HTML}{FEFEFE}
\definecolor{arctic10}{HTML}{68625F}
\definecolor{semiarid0}{HTML}{898A79}
\definecolor{semiarid1}{HTML}{655745}
\definecolor{semiarid2}{HTML}{343330}
\definecolor{semiarid3}{HTML}{D6C19D}
\definecolor{semiarid4}{HTML}{D6C19D}
\definecolor{sahara0}{HTML}{D9A27B}
\definecolor{sahara1}{HTML}{95613E}
\definecolor{sahara2}{HTML}{895332}
\definecolor{sahara3}{HTML}{9D6C4B}
\definecolor{sahara4}{HTML}{663415}
\definecolor{sahara5}{HTML}{794424}
\definecolor{sahara6}{HTML}{522607}
\definecolor{sahara7}{HTML}{AB7B60}
\definecolor{urban10}{HTML}{323B36}
\definecolor{urban11}{HTML}{4D524D}
\definecolor{urban12}{HTML}{B0A7A2}
\definecolor{urban13}{HTML}{4D524D}
\definecolor{urban14}{HTML}{EFE5E2}
\definecolor{urban15}{HTML}{95140F}
\definecolor{urban16}{HTML}{6D513B}
\definecolor{urban17}{HTML}{87979B}
\definecolor{urban20}{HTML}{7C6140}
\definecolor{urban21}{HTML}{95835E}
\definecolor{urban22}{HTML}{13110F}
\definecolor{urban23}{HTML}{13110F}
\definecolor{urban24}{HTML}{BFB18B}
\definecolor{urban25}{HTML}{F0ECDD}
\definecolor{urban26}{HTML}{463E3D}
\definecolor{urban27}{HTML}{46251F}
\definecolor{urban30}{HTML}{928F86}
\definecolor{urban31}{HTML}{7A7872}
\definecolor{urban32}{HTML}{67655E}
\definecolor{urban33}{HTML}{504E46}
\definecolor{urban34}{HTML}{A8A8A2}
\definecolor{urban35}{HTML}{38362E}
\definecolor{urban36}{HTML}{1A1A17}
\definecolor{urban37}{HTML}{D1CEC8}
\definecolor{night10}{HTML}{141213}
\definecolor{night11}{HTML}{2C2730}
\definecolor{night12}{HTML}{312A40}
\definecolor{night13}{HTML}{586BA3}
\definecolor{night14}{HTML}{4672A6}
\definecolor{night15}{HTML}{202006}
\definecolor{night16}{HTML}{9B9B3A}
\definecolor{night17}{HTML}{CBCBA5}
\definecolor{night20}{HTML}{152418}
\definecolor{night21}{HTML}{414741}
\definecolor{night22}{HTML}{302D2C}
\definecolor{night23}{HTML}{493F39}
\definecolor{night24}{HTML}{78746B}
\definecolor{night30}{HTML}{1C1618}
\definecolor{night31}{HTML}{120F02}
\definecolor{night32}{HTML}{72B09B}
\definecolor{night33}{HTML}{753FF2}
\newcommand{\orcid}[1]{\href{https://orcid.org/#1}{\includegraphics[scale=0.06]{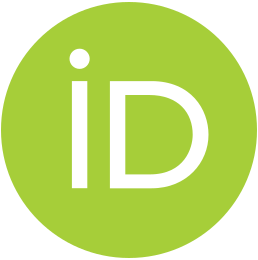}}}

\title{Structured Adversarial Camouflage via Voronoi Diagrams}

\author{
        Jens Bayer$^{1,2 \text{ \orcid{0000-0002-2806-6920}
        \href{mailto:jens.bayer@iosb.fraunhofer.de}{\tiny\faEnvelope}}}$
        \And
Stefan Becker$^{1 \text{ \orcid{0000-0001-7367-2519}}}$\And
David M\"unch$^{1 \text{ \orcid{0000-0002-8577-5256}}}$\And
Michael Arens$^{1 \text{ \orcid{0000-0002-7857-0332}}}$\And
J\"urgen Beyerer$^{1,2 \text{ \orcid{0000-0003-3556-7181}}}$\And
        \\
    $^1$ Fraunhofer IOSB and Fraunhofer Center for Machine Learning,\\
    $^2$ Karlsruhe Institute of Technology
}

\renewcommand{\shorttitle}{Structured adv. Camouflage via Voronoi Diagrams}

\hypersetup{
pdftitle={Structured Adversarial Camouflage via Voronoi Diagrams},
pdfauthor={Jens Bayer, Stefan Becker, David M\"unch, Michael Arens, J\"urgen Beyerer},
pdfkeywords={adversarial attacks, camouflage, adversarial pattern},
}
\begin{document}
\maketitle

\begin{abstract} 

Pixel-wise adversarial patches are computationally heavy and often visually
	detectable, limiting utility in security-critical systems. We present
	adversarial Voronoi camouflage that optimizes only seed-point locations
	under fixed, printable palettes using a soft assignment, producing
	structured, splinter camouflage-like patterns without additional
	regularization. Evaluated on person detection with COCO-style
	AP@[.5:.95], naive placement (Inria $\rightarrow$ COCO) performs
	comparable bad, while garment-level application via segmentation mask
	(3DPeople) results in a significant AP drop. The attack transfers to
	out-of-domain backgrounds and across detector families
	(YOLOv9/10/11/12), indicating robustness in black-box settings.
	Repainting with different palettes largely nullifies the effect, and
	single-color tweaks show limited tolerance ($\leq$0.17), highlighting a
	structure–palette coupling. The parameter-efficient,
	palette-constrained design improves visual plausibility while degrading
	real-time detector performance. Physical validation and color
	calibration are left for future work.\\
	Code: \url{https://github.com/JensBayer/Voronoi}\\\\
	This paper was originally presented at the International Conference on
	Military Communication and Information Systems (ICMCIS), organized by
	the Information Systems Technology (IST) Scientific and Technical
	Committee, IST-224-RSY – the ICMCIS, held in Bath, United Kingdom,
	12-13 May 2026.
\end{abstract}

\section{Introduction} 
Modern reconnaissance and surveillance architectures in defense increasingly
rely on deep-learning-based object detectors deployed on unmanned platforms,
mobile systems, and fixed sensor networks. They often form the first stage of
a chain, informing human operators or downstream decision components to the
presence of personnel or materiel. Any systematic weakness of these models
can therefore have a disproportionate operational impact. Prior work has shown
that carefully crafted physical patterns on clothing, vehicles, or equipment can
significantly reduce detection performance under realistic viewing
conditions~\cite{Chakraborty2021,Wei2024,Wu2020,Xu2020,Duan2020,Duan2021,Zhu2022},
making adversarial camouflage both a potential capability for friendly forces
and a threat if exploited by adversaries. Nonetheless, many existing methods
produce visually conspicuous patterns, deviate strongly from established
camouflage styles, or rely on color choices that are difficult to realize with
real pigments and textiles, which hampers technology transfer and robust
testing.

The adversarial Voronoi patterns proposed in this work directly address these
issues by providing a structured, palette-constrained parameterization that is
much closer to current camouflage design practice. By optimizing only seed-point
positions using fixed color palettes, we obtain splinter-like patterns that
resemble naturalistic woodland, urban, or desert designs, yet measurably
suppress person detections across a range of modern real-time detectors and
backgrounds. This makes the approach attractive both as a building block for
future physical decoys and as a digital red-team tool for defense stakeholders
to probe the vulnerability of their detection pipelines. It also serves to
inform the development of hardening strategies such as adversarial training or
tailored data augmentation.

\begin{figure}[tb]
    \centering
    \includegraphics[width=0.3\textwidth]{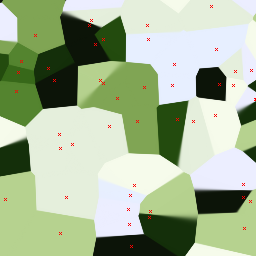}
    \caption{An optimized Voronoi patch with the corresponding seed points (red
    crosses). Due to the soft assignment, some cells are separated less
    sharply.}
    \label{fig:optimization-process}
\end{figure}

In summary, this work presents (1) a novel approach to generating adversarial
patterns using Voronoi diagrams (see \autoref{fig:optimization-process}), with a
fixed, predefined color-palette of choice. The extensive evaluation covers a (2)
garment-level application using segmentation masks
(3DPeople~\cite{Pumarola2019}) that verifies the effectiveness as a potential
camouflage cloth pattern and (3) transferability across multiple backgrounds
(BG-20k~\cite{Li2022}) as well as (4) transferability across YOLOv9/10/11/12.
Moreover, an (5) analysis of color palette tolerance, including practical
implications for real-world physical implementation, is given. We explicitly do
not claim physical stealth; validating printability, color calibration, garment
deformation, and human factors requires controlled, ethics-approved studies.

\section{Related work}
As the topic of adversarial attacks and robustness has gained increased interest
recently, the amount of related work has also grown significantly. For a broader
overview of adversarial attacks, we refer to~\cite{Chakraborty2021}
and~\cite{Wei2024} as these surveys cover related work that exceeds the topic of
camouflaged adversarial attacks.

Wu et al.~\cite{Wu2020} systematically investigate the transferability of
adversarial attacks on different object detection frameworks. By training
adversarial patches designed to suppress objectness scores, they demonstrate
that these attacks can effectively transfer across various detectors and
datasets, both digitally and physically. Their exploration of physical-world
attacks using printed materials highlights the practical implications of such
vulnerabilities. This aspect bridges the gap between theoretical findings and
real-world applications, emphasizing the potential risks of adversarial attacks
in the physical world.
A further refinement of the generation of physical-world attacks is introduced
by Xu et al.~\cite{Xu2020} with their \emph{adversarial T-shirts}. To tackle the
problems of possible cloth deformations caused by a moving, non-rigid object,
they use thin plate spline mapping during the optimization of the pattern.

Regarding adversarial camouflage patterns that are not visible to a human
observer, Duan et al.~\cite{Duan2020} introduce \emph{AdvCam}. A highly stealthy yet
effective method to hide adversarial patterns in the digital and physical
world. The generated patterns successfully attack multiple classifiers.

\emph{Dense Proposals Attack} (DPA)~\cite{Duan2021} further advances the field by
generating one-piece adversarial camouflages designed to be robust under various
viewpoints and lighting conditions. The successful implementation of DPA on a 3D
printed vehicle demonstrates its practical applicability. However, it is worth
noting that the current camouflages generated by DPA have suspicious
appearances, which could limit their effectiveness in real-world deployments.

The research has also extended these concepts into the domain of thermal
infrared imaging with the development of infrared adversarial clothing using
aerogel material~\cite{Zhu2022}. Zhu et al. optimize binary patterns that look
similar to QR codes and can fool thermal infrared object detectors, even
when worn by people moving dynamically.

An integration of military camouflage principles into patch-based attacks is
presented by \cite{Kim2023a}. The authors address the high human perceptibility
by proposing a method that enhances the naturalistic appearance of the patches
while maintaining their effectiveness in fooling object detectors.

Wang et al.~\cite{Wang2023a} combine findings from Xu et al.~\cite{Xu2020} and
Duan et al.~\cite{Duan2020} to generate \emph{Dual Attribute Adversarial
Camouflage} (DAAC) for evading detection by both detectors and humans. They
extract scene features from the set of a specific scene, including dominant
colors and proportions, and uses these to generate digital camouflage and
constrain the adversarial patch during training. The authors acknowledge
challenges in implementing DAAC in practice, such as material constraints and
environmental variations.

Most similar to our method is~\cite{Peng2025}, where the authors use
StyleGAN3~\cite{karras2021alias} as a pretrained texture generator to produce
camouflage pattern. Once trained, the latent input variables of the generator
are optimized in an adversarial fashion, causing a significant impact on the
detector performance.

Recent work by Van der Burg et al.~\cite{VanderBurg2025} has explored the
intersection of automated evaluation and human perception in camouflage
effectiveness. They demonstrated that pre-trained CNNs, such as YOLO, can serve
as proxies for human detection capabilities, providing an image-based measure of
camouflage success. However, discrepancies emerged at close distances, where
high-contrast or shape-breaking elements in the camouflage patterns
significantly affected both human and model performance.

Building upon these foundations, this work introduces \emph{adversarial Voronoi
patches}. Instead of optimizing each pixel individually, we leverage the
geometric structure of Voronoi diagrams by optimizing seed points with a fixed
color palette. This approach offers several advantages over traditional
pixelwise optimization techniques, e.g. inherently more structured adversarial
patterns that are less likely to be detected by human observers, addressing the
visual perceptibility limitations observed in previous studies.

\section{Adversarial Voronoi Patches}
We define a differentiable Voronoi pattern as follows: Given a set of seed
points $\mSet{S} = \{s_1,\ldots,s_n\}\subset \mathbb{R}^{2}$, and a pixel grid $\mSet{I}
= \{1,\ldots,H\} \times \{1,\ldots,W\}$. Assign each seed a color $\mVector{c}_i\in
[0,1]^3$. For $\mVector{p} \in \mSet{I}$ define distances
$\mFunction{d}_i(\mVector{p}) = \|\mVector{p} - s_i\|$. 
We obtain a temperature-scaled softmin over distances 
\begin{equation}
	w_i(\mVector{p}, \tau) = \frac{e^{\frac{-d_i(\mVector{p})}{\tau}}}{\sum_{k=1}^{n} e^{\frac{-d_k(\mVector{p})}{\tau}}}.
\end{equation}
As $\tau \rightarrow 0$ this approaches a one-hot at $\mFunction{arg\,min}_i\, d_i(\mVector{p})$

The final Voronoi pattern is the convex combination
\begin{equation}
	\mFunction{f}(\mVector{p}) = \sum_{i=1}^{n} w_i(\mVector{p},\tau) \cdot \mVector{c}_i.
\end{equation}
After the pattern is generated, it is placed onto objects of interest, and
the resulting image is propagated through an object detector. The prediction of
the detector is then used to minimize the detector's confidence for classes of
interest present in the given image, resulting in the optimization of the seed
points.


\section{Experimental Setup}
In total, five experiments are conducted, where each uses \enquote{Person} as
the target class for the patches. The weights of the investigated detectors are
the official pretrained weights, optimized with the COCO~\cite{Lin2014} dataset.

Experiment 1 is a general investigation of whether adversarial Voronoi
patches can be optimized to fool object detectors and how well they perform when
optimized like regular adversarial patches. Since regular adversarial patches
are allowed to freely optimize pixels, the expectation is that Voronoi
patches perform comparatively poorly.
For experiment 2, the performance of the patches is investigated when
the optimization is performed using segmentation masks during training. The
segmentation masks are used to replace the cloth of a depicted person with parts
of the adversarial patch. As the segmented clothes cover a bigger part of a
person, the patch should perform much better.
The transferability of the generated patches regarding different object detectors
is evaluated in experiment 3. Moreover, the segmentation information is
used to place the person instances on different, out-of-domain backgrounds.
Experiment 4 checks the transferability of the adversarially optimized structure
by exchanging the color palettes of the trained patch set of experiment 2.
In the final experiment 5, the generated patches from the first and second
experiments are exchanged to investigate whether the different optimization
methods result in different patch performances.

All generated patterns have 256 seed points, are randomly initialized in
$[0,1]^2$, and use the temperature parameter $\tau=0.001$. The width and
height of the patterns are set to 256 pixels.

\subsection{Experiment 1: Naive Patches}
The first experiment evaluates the performance of an adversarial Voronoi patch
optimized in the same way as a regular adversarial patch is trained. During the
optimization of a patch, it is naively placed inside the bounding box of
objects of interest, and the detector's \enquote{objectness score} is
successively minimized. Since more recent YOLO variants no longer calculate
objectness scores, the global maximum pre-logit class scores of both detection
heads are selected instead. The augmentation pipeline includes a random resize
$[0.6,0.9]$, random rotation $[-\ang{30},\ang{30}]$, color jitter, and random
perspective. While a regular adversarial patch has two additional loss terms
that punish the use of invalid pixel colors and smoothen the patch, the
adversarial Voronoi patch has no such regularization terms.
The experiment uses the InriaPerson~\cite{Dalal2005} dataset for training and the
patches and the COCO~\cite{Lin2014} dataset for the evaluation. While the former
is a pure person dataset, the latter is a well-known benchmark dataset
containing a vast amount of person instances.
Patches are optimized over 100 epochs using the AdamW~\cite{Loshchilov2019}
optimizer with an initial learning rate of 0.01 and a reduction of the learning
rate by a factor of 10 every 25 epochs.

\subsection{Experiment 2: Covering Clothes}
\begin{figure}[tb]
    \centering
    \includegraphics[width=0.35\linewidth]{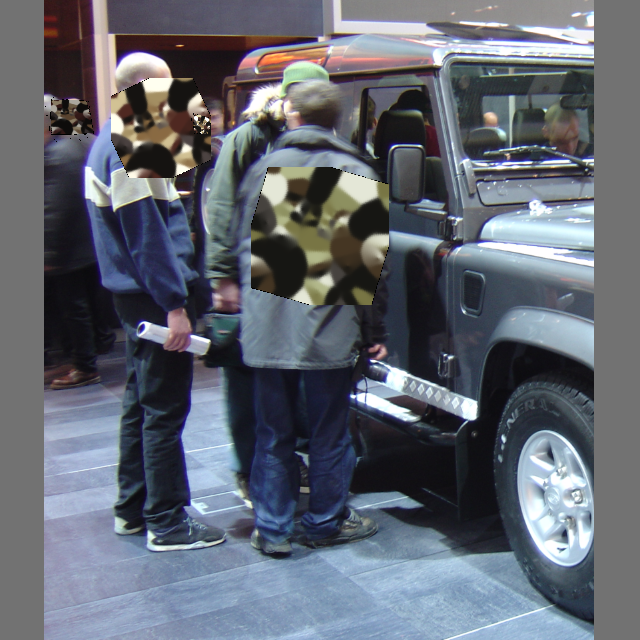}
    \includegraphics[width=0.35\linewidth]{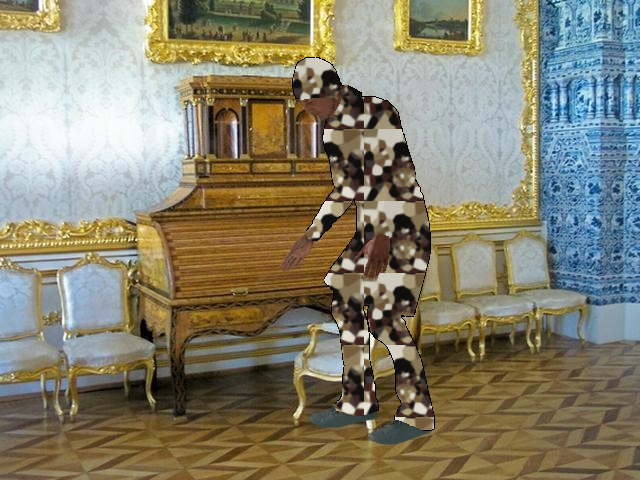}
    \caption{Example image of the InriaPerson dataset and the 3DPeople dataset.
    Due to the lack of cloth segmentation masks, the patches optimized with the
    InriaPerson dataset and COCO dataset are simply placed on objects of
    interest. For 3DPeople, the segmentation mask is used to cover the
    clothed parts of the person with the adversarial patch.}
    \label{fig:person3d}
\end{figure}

In a subsequent experiment, the 3DPeople dataset~\cite{Pumarola2019} is used to
optimize Voronoi patches. This dataset includes synthetic 3D renderings of
people in varied environments and provides the segmentation masks for each
instance's clothing. These masks are utilized to refine adversarial Voronoi
patches, ensuring they cover the clothed regions of individuals instead of just
placing them inside the bounding box of objects of interest (see
\autoref{fig:person3d}).
Since the amount of images in the training set is larger, as in experiment 1,
patches are optimized over 10 epochs. Again, AdamW is used with a learning rate
of 0.001 and a reduction of the learning rate every three epochs by a factor of
10.

\subsection{Experiment 3: Transferability}
The third set of experiments examines the transferability of the generated
patches of experiment 2 when they are used to attack other object detectors and
different background images.
For the change of the background image, the segmentation masks provided by the
3DPeople dataset are used to cut out the person instances and place them onto
center-cropped images of the BG-20k dataset~\cite{Li2022}. To reduce the
cut-out effect, a small area around the border of the person and the background
image is blurred with a Gaussian kernel.
The transferability study conducted in this experiment should be interpreted
with caution, as solely YOLOv10b is used to optimize the patches.

\subsection{Experiment 4: Changes in the color palette}
Since adversarial Voronoi patches are optimized using a fixed color palette, the
palettes can be exchanged easily. The resulting patch retains the structure but
changes the appearance due to the different color. The fourth experiment answers
whether the optimized structure can be combined with a different color palette
and thus also retain a degree of the adversarial property.

\subsection{Experiment 5: Naive Patches for Clothes and Vice Versa}
In a final experiment, the trained patches of experiment 1 and experiment 2
are exchanged and evaluated in the corresponding different setting.

\subsection{Real-time Object Detector}
As YOLO detectors are a well-known and investigated real-time object detector
family, the YOLOv10 detector is selected as the detector to be attacked. As
shown by Bayer et al.~\cite{Bayer2024}, the transferability of a patch generated
with the YOLOv10 architecture provides good transferability across various
architectures. Moreover, the lightweight and fast YOLOv10b variant allows a
faster and thus a more comprehensive evaluation. The missing non-maximum
suppression in YOLOv10 results in the use of two detector heads during the
detector training. While the one-to-many head is usually disabled during
inference, the outputs of both of these heads are used during patch
optimization.


\subsection{Color Palettes}
\begin{figure}[tb]
    \centering
    \begin{tikzpicture}
    \node [anchor=east] at (-0.5,0) {Forest Spring:};
	\foreach \x in {0,...,10}{
		\filldraw[fill=forestspring\x](\x*0.5,0) circle (0.2);
	};
    \node [anchor=east] at (-0.5,-0.5) {Coastal:};
	\foreach \x in {0,...,10}{
		\filldraw[fill=coastal\x](\x*0.5,-0.5) circle (0.2);
	};
    \node [anchor=east] at (-0.5,-1) {Sahara:};
	\foreach \x in {0,...,7}{
		\filldraw[fill=sahara\x](\x*0.5,-1) circle (0.2);
	};
    \node [anchor=east] at (-0.5,-1.5) {Urban 1:};
	\foreach \x in {0,...,7}{
		\filldraw[fill=urban1\x](\x*0.5,-1.5) circle (0.2);
	};
    \node [anchor=east] at (-0.5,-2) {Night 1:};
	\foreach \x in {0,...,7}{
		\filldraw[fill=night1\x](\x*0.5,-2) circle (0.2);
	};
    \end{tikzpicture}
    \caption{Five of the thirteen available color palettes that are used for the
    evaluation. The color values are adopted from Camogen \cite{Camogen2018}.}
    \label{fig:colors}
\end{figure}
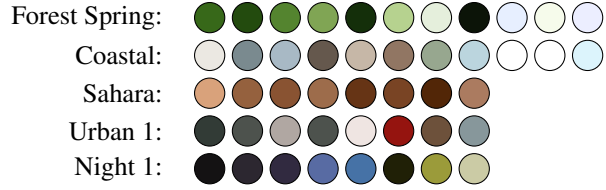

The color palettes used for the evaluation are adopted from \cite{Camogen2018}.
Of the 13 available palettes, a total of 5 are evaluated. The selected palettes
are given in \autoref{fig:colors} and are randomly chosen. For each palette,
three random Voronoi patches are generated that contain only colors of the
respective palette. These patches are used in the different experiments as 
points of reference.

\section{Evaluation}
The experiments use the average precision (AP) with multiple intersection over
union thresholds (0.5 to 0.95 in steps of 0.05) to measure the change of the
detector's performance.
In all following tables, $\uparrow (\downarrow)$ indicates whether a higher
(lower) value is desired. $\Delta$ AP is the relative performance of a patch
and is the difference between the achieved AP in the specific setting and the
AP achieved with the reference patches in the same setting. If not stated
otherwise, a reference patch shares the same parameterization (number of cells,
color palette) but is not optimized.

\subsection{Experiment 1: Naive Patches}
The expectation of the performance of adversarial Voronoi patches was met: in
comparison to a regular adversarial patch, the damping of the patch performance
compared to the reference patches is low (see \autoref{tab:experiment1}). Yet,
the lowering of the AP as a result of the optimization process affirms that
Voronoi patches can contain an adversarial component. 
Surprisingly, the reference patches of all palettes result in a similar AP that
differs after the third decimal place.
The usage of the \emph{Sahara} palette, which is also the palette with the least
variant colors, results in patches that have the strongest adversarial
component. The least adversarial effect is achieved with the \emph{Forest
Spring} palette.
\begin{table}[tb]
    \centering
    \caption{Experiment 1: Patches are optimized with InriaPerson and evaluated
    on COCO. The reference AP is the performance of the detector when random
    Voronoi patterns of the same palette without any seed point optimization are
    used.}
    \label{tab:experiment1}
    \begin{tabular}{c c c c}
    \toprule
    Palette & ref. AP & AP $\downarrow$ & $\Delta$ AP $\uparrow$ \\
    \midrule
    Forest Spring & $0.45\pm0.00$ & $0.42\pm0.01$ & $0.02\pm0.01$\\
    Coastal & $0.45\pm0.00$ & $0.41\pm0.00$ & $0.04\pm0.00$\\
    Sahara & $0.45\pm0.00$ & $\mathbf{0.39\pm0.01}$ & $\mathbf{0.06\pm0.01}$\\
    Urban 1 & $0.45\pm0.00$ & $0.42\pm0.01$ & $0.03\pm0.01$\\
    Night 1 & $0.45\pm0.00$ & $0.41\pm0.01$ & $0.04\pm0.01$\\
    \bottomrule
    \end{tabular}
    \end{table}

\subsection{Experiment 2: Covering Clothes}
When the dataset and optimization strategy are changed, the performance of both
the reference patches and optimized patches changes drastically (see
\autoref{tab:experiment2}). The performance of the object detector when the
reference patches are applied on the 3DPeople Dataset is much better compared
to the COCO case. This could be due to the less disruptive application of the
patches: For the 3DPeople dataset, the patch does not cover edges of objects of
interest but is rather part of the object itself (see \autoref{fig:person3d}).
In experiment 1, the patches are applied more invasively. In addition, the
adversarial structure covers a larger area of the person.
The highest AP drop is recorded for \emph{Night 1}, where the reference AP
dropped from $0.70\pm0.01$ to $0.26\pm0.03$. Furthermore, the AP of patches of
all other palettes dropped significantly by at least $0.32$.
Despite being able to detect persons \enquote{wearing} the reference patches
easily, the detector performance breaks down when optimized patches are
\enquote{worn}. This finding indicates the containment of an adversarial
component in the presented Voronoi patches, thus making the presented system a
valid consideration for future camouflage patterns.

\begin{table}[tb]
    \centering
    \caption{Experiment 2: The reference AP is the performance of the detector
    when receiving a random Voronoi pattern of the same palette without any
    seed~point optimization. The evaluation is performed using the test split
    with the original background images of the 3DPeople Dataset.}
    \label{tab:experiment2}
    \begin{tabular}{c c c c}
    \toprule
    Palette & ref. AP & AP $\downarrow$ & $\Delta$ AP $\uparrow$ \\
    \midrule
    Forest Spring & $0.60\pm0.01$ & $0.29\pm0.01$ & $0.32\pm0.01$\\
    Coastal & $0.59\pm0.03$ & $0.26\pm0.03$ & $0.33\pm0.00$\\
    Sahara & $0.61\pm0.01$ & $0.26\pm0.02$ & $0.34\pm0.02$\\
    Urban 1 & $0.65\pm0.04$ & $0.29\pm0.02$ & $0.36\pm0.03$\\
    Night 1 & $0.70\pm0.01$ & $\mathbf{0.26\pm0.03}$ & $\mathbf{0.43\pm0.03}$\\
    \bottomrule
    \end{tabular}
\end{table}

\subsection{Experiment 3: Transferability}
\subsubsection{Background Transferability}
\autoref{tab:experiment3a} shows the transferability capability of the generated
patches of experiment 2 when using different background sceneries for the
3DPeople dataset. Compared to the results of experiment 2, the changes of the
backgrounds improve the AP for the reference patterns. A probable cause is the
strong border that results from the cut-out process with the segmentation mask.
The highest AP drop of $0.51$ is again recorded for \emph{Night 1}, followed by
\emph{Urban 1} ($0.48$) and \emph{Coastal} ($0.45$).

\begin{table}[tb]
    \centering
    \caption{Experiment 3.1: Background images are changed to images of
    the BG-20k dataset. The reference AP is the performance of the
    detector when receiving a random Voronoi pattern of the same palette without
    any seed~point optimization.}
    \label{tab:experiment3a}
    \begin{tabular}{c c c c }
        \toprule
        Palette & ref. AP & AP $\downarrow$ & $\Delta$ AP $\uparrow$\\
        \midrule
        Forest Spring& $0.86\pm0.01$ & $0.47\pm0.05$ & $0.39\pm0.00$\\
        Coastal      & $0.88\pm0.01$ & $0.43\pm0.02$ & $0.45\pm0.00$\\
        Sahara       & $0.86\pm0.02$ & $0.51\pm0.01$ & $0.35\pm0.00$\\
        Urban 1      & $0.88\pm0.01$ & $0.40\pm0.03$ & $0.48\pm0.00$\\
        Night 1      & $0.89\pm0.01$ & $\mathbf{0.38\pm0.02}$ & $\mathbf{0.51\pm0.00}$\\
        \bottomrule
    \end{tabular}
\end{table}

\subsubsection{Model Transferability}
\begin{figure}[tb]
    \centering
    \includegraphics[width=0.5\linewidth]{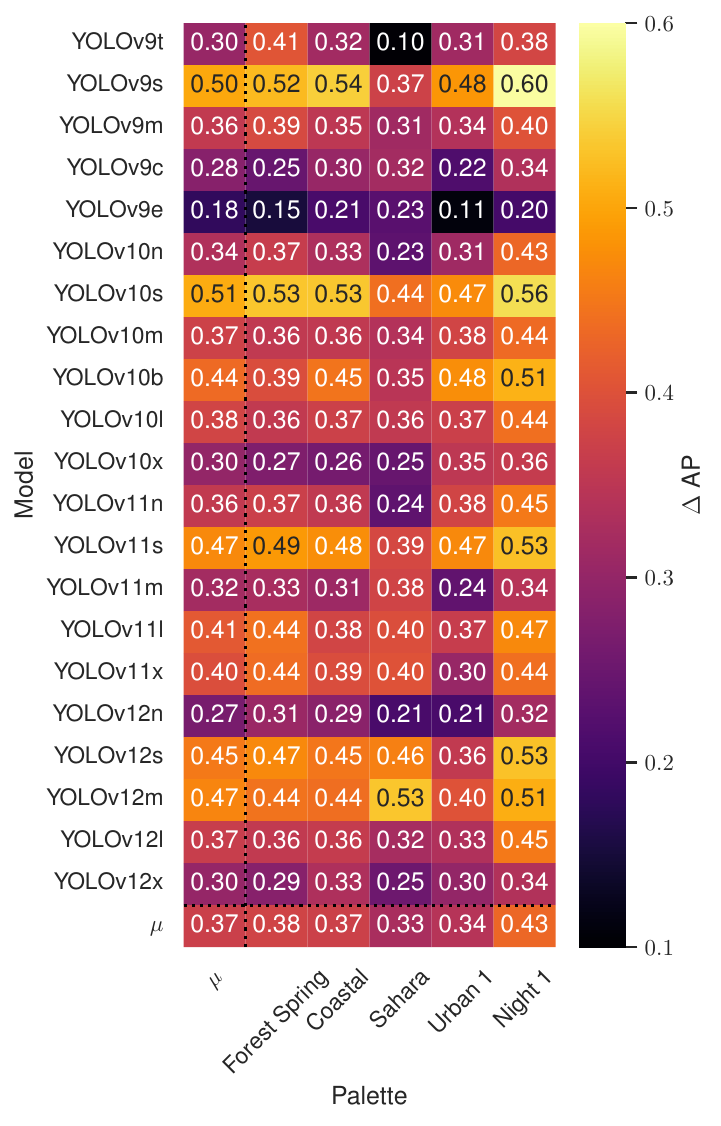}
    \caption{Experiment 3.2: The transferability matrix of different network
    architectures and the five selected palettes shows the impact of adversarial
    Voronoi patches in a black-box setting. The lighter the shade of a cell, the
    higher the AP drop is. The AP difference is calculated in respect to a set
    of three random initialized patterns of the respective palette. $\mu$
    indicates the respective column and row mean value.}
    \label{fig:experiment3b}
\end{figure}

The model transferability for different detector architectures is shown in
\autoref{fig:experiment3b}. The given transferability matrix evaluates different
YOLO architectures (y-axis) with the optimized patches of experiment 2 of the
five selected palettes (x-axis). Each cell of the matrix is shaded according to
the relative AP drop in respect to the performance the model achieves when
evaluated with the respective reference patterns. The first column and last row
are labeled with $\mu$ and are the row-wise and column-wise means.
The overall mean performance drop is about $0.37$. The lowest mean performance
drops are given by \emph{Sahara} ($0.33$) and \emph{Urban 1} ($0.34$) while
\emph{Night 1} achieves the highest ($0.43$) AP drop. The most robust model is
YOLOv9e, with a mean AP drop of $0.18$ followed by YOLOv12n ($0.27$). The least
robust ones are YOLOv10s ($0.51$) and YOLOv9s ($0.50$). Other than expected, the
performance drop of some smaller models is far less than the drop of some larger
models. This counterintuitive observation can be explained by taking a look at
the absolute values: the performance on the reference pattern
differs for the smallest models of an architecture group by about 20 AP points
in comparison to the largest. Consequently, the overall performance when attacked
is also much lower.

\subsection{Experiment 4: Changes in the color palette}
\subsubsection{Change the palette as a whole}
\begin{figure}[tb]
    \centering
    \includegraphics[width=0.4\linewidth]{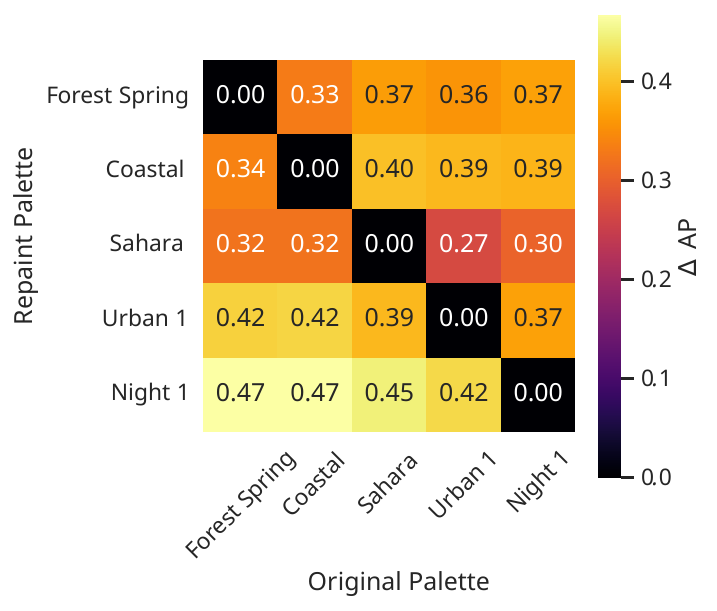}
    \caption{Experiment 4.1: Patches are optimized with the palette at the
    x-axis and evaluated with colors of the palette given in the y-axis. The
    color of a cell represents the difference between the AP of the patch when
    the original palette is used and the AP when the respective repaint palette
    is used. The darker the shade, the better the set of patches perform.}
    \label{fig:experiment4a}
\end{figure}

Simply exchanging a single color palette with another palette for the patches
trained in experiment 2 results in a significant loss of performance, as
depicted in \autoref{fig:experiment4a}. Here, each row and column represents a
single color palette. The palettes on the x-axis are the base palettes used to
train the pattern. The different palettes used to exchange the colors are
plotted on the y-axis. Again, the shade of a cell represents the change of the
AP of the patches, but this time regarding the respective performance when
the original palette is used to generate the patch.
The main diagonal is equal to zero, as these patches will always achieve the
same performance with their respective original palettes. By exchanging the
colors with another palette, all patches lose their adversarial component and
perform after the color swap similar to the set of reference patches.

\subsubsection{Impact of changes in a single palette color}
\begin{figure*}[tb]
    \centering
    \includegraphics[width=0.3\linewidth]{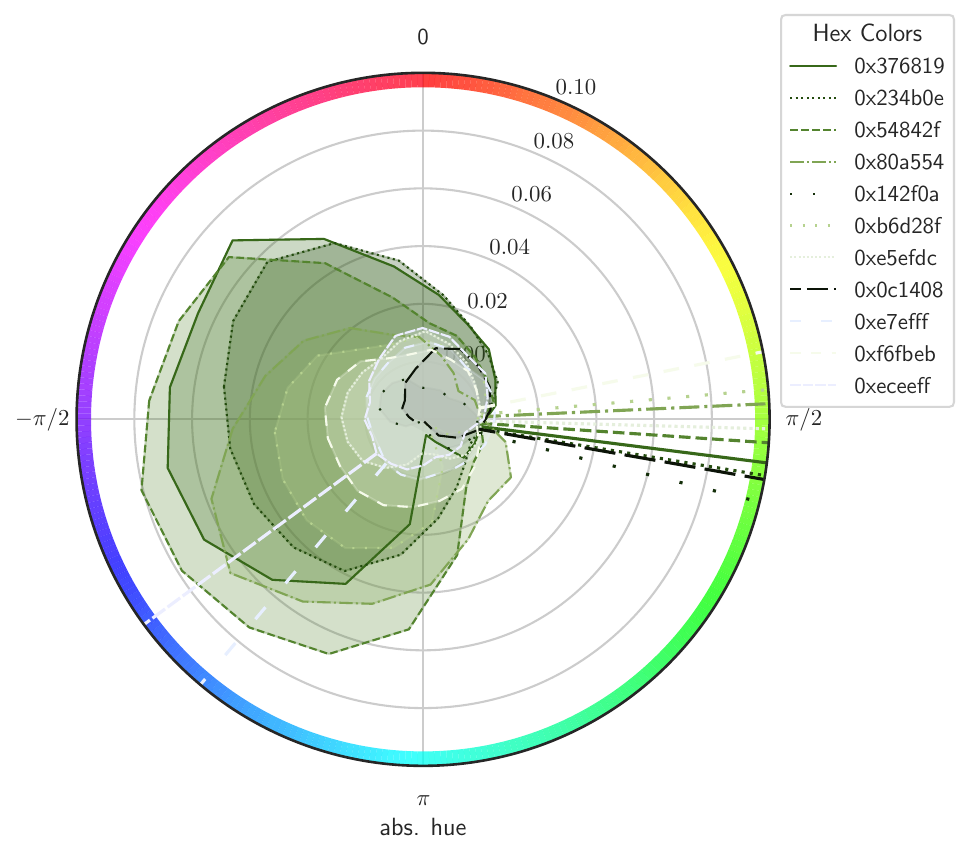}
    \includegraphics[width=0.3\linewidth]{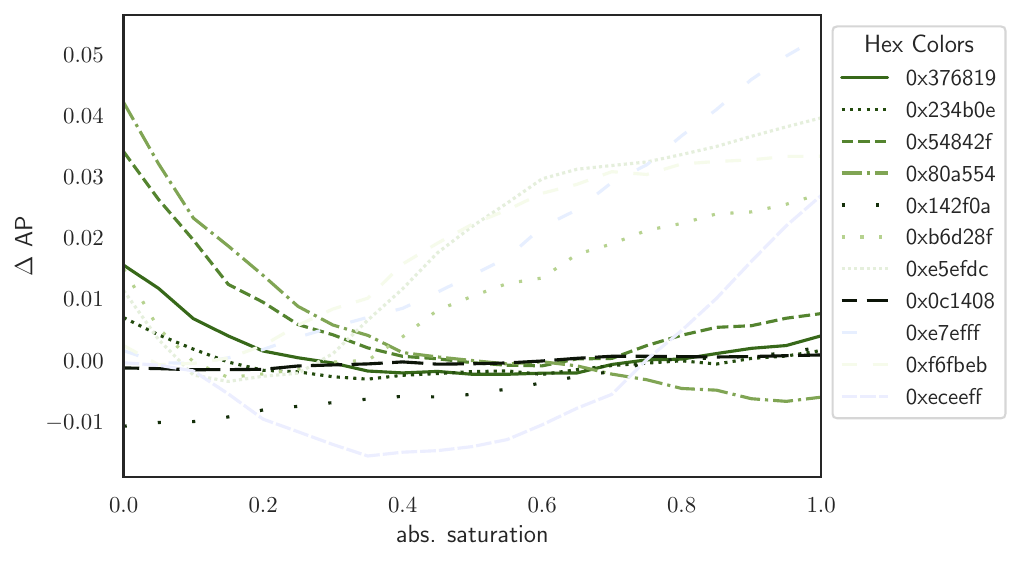}
    \includegraphics[width=0.3\linewidth]{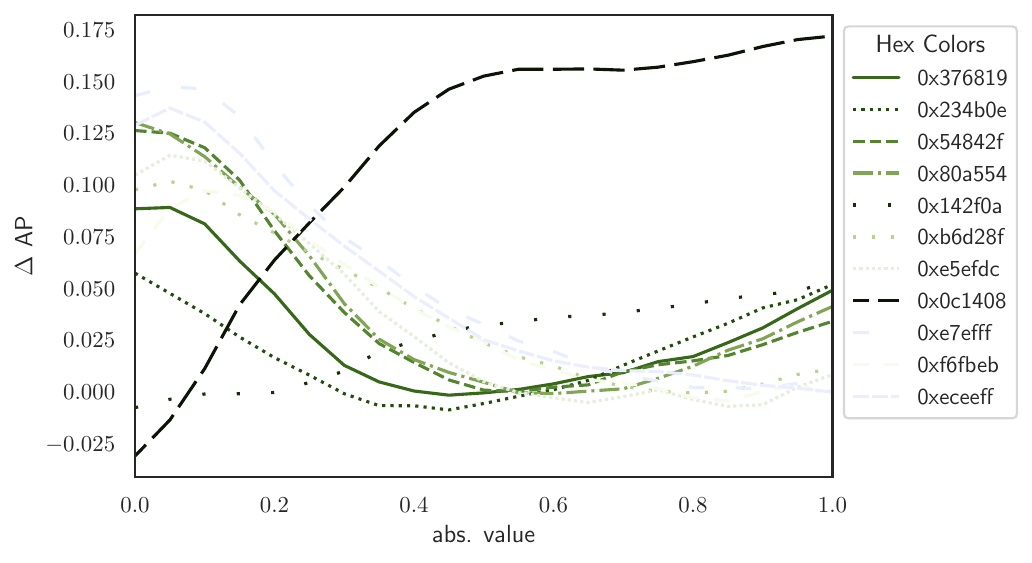}
    \caption{Experiment 4.2.: Impact on the AP for changes in hue, value, and
    saturation using the forest spring palette on one of the previous
    optimized patches.}
    \label{fig:experiment4b}
\end{figure*}

Slightly changing a single color of a palette while fixing all the other colors
is not as impactful as changing the whole palette. \autoref{fig:experiment4b} contains
three columns, where each column visualizes the impact of changes in hue, value,
and saturation of colors of a palette given a specific patch.
The radar plot depicts the hue space. The radius to the center point represents
the relative change in the AP. Each colored area represents a single color of
the palette. The straight line that shares the line style and color as an area
represents the unaltered hue value of that color. For example, when following
the straight green line clockwise (0x376819), one observes that a change of the
color to a cyan color (at around $\pi$) would increase the adversarial component
of the pattern. If the hue value is shifted even more towards purple, the patch
successively loses its adversarial component. This plot gives a hint on how high the
tolerance should be if a pattern is to be realized in the physical world.
The central plot provides information on how the shift in saturation of a certain
color changes the performance of the patch. For example, the striped, dark
green line (0x0c1408) barely changes as saturation increases. In the rightmost
plot, where the influence of the value component is depicted, the dark green
line (0x0c1408) behaves entirely differently. The change in the value lightens
the color up to an almost lime green hue, and the patch AP increases about
$0.17$ AP points.

\subsection{Experiment 5}
The last experiment evaluates the patches trained in experiment 1 the same way
the patches of experiments 2 and 3 are evaluated, and vice versa. 

When comparing \autoref{tab:experiment5a} with \autoref{tab:experiment1}, the
performance of the 3DPeople patches is slightly better than the patches
optimized in experiment 1. Nonetheless, the overall performance is still low.
This is also true for the performance of the patches that are optimized with
InriaPerson and evaluated with the 3DPeople dataset (see
\autoref{tab:experiment5b}). Unfortunately, the AP drop is much lower than the
patches optimized on the 3DPeople dataset (see \autoref{tab:experiment3a}). This
could be due the reason, that InriaPerson contain much less images compared to
3DPeople and thus can offer much less variation during the patch optimization.
\begin{table}[tb]
    \centering
    \caption{Experiment 5.1: 3D People Patches evaluated with COCO. The reference
    AP is again the performance of the detector when receiving a random Voronoi
    pattern of the same palette without any seed~point optimization.}
    \label{tab:experiment5a}
    \begin{tabular}{c c c c}
    \toprule
    Palette & ref. AP & AP $\downarrow$ & $\Delta$ AP $\uparrow$ \\
    \midrule
    Forest Spring & $0.45\pm0.00$ & $\mathbf{0.38\pm0.02}$ & $\mathbf{0.07\pm0.02}$\\
    Coastal & $0.45\pm0.00$ & $0.39\pm0.01$ & $0.06\pm0.01$\\
    Sahara & $0.45\pm0.00$ & $0.41\pm0.01$ & $0.04\pm0.01$\\
    Urban 1 & $0.45\pm0.00$ & $0.40\pm0.01$ & $0.05\pm0.01$\\
    Night 1 & $0.45\pm0.00$ & $0.41\pm0.01$ & $0.04\pm0.01$\\
    \bottomrule
    \end{tabular}
\end{table}

\begin{table}[tb]
    \centering
    \caption{Experiment 5.2: InriaPerson Patches evaluated with the 3DPeople dataset. The reference
    AP is again the performance of the detector when receiving a random Voronoi
    pattern of the same palette without any seed~point optimization.}
    \label{tab:experiment5b}
    \begin{tabular}{c c c c}
    \toprule
    Palette & ref. AP & AP $\downarrow$ & $\Delta$ AP $\uparrow$ \\
    \midrule
    Forest Spring & $0.86\pm0.01$ & $0.80\pm0.02$ & $0.07\pm0.02$\\
    Coastal & $0.88\pm0.01$ & $\mathbf{0.79\pm0.01}$ & $\mathbf{0.08\pm0.01}$\\
    Sahara & $0.86\pm0.02$ & $0.79\pm0.02$ & $0.06\pm0.02$\\
    Urban 1 & $0.88\pm0.01$ & $0.80\pm0.03$ & $0.07\pm0.03$\\
    Night 1 & $0.89\pm0.01$ & $0.81\pm0.01$ & $0.08\pm0.01$\\
    \bottomrule
    \end{tabular}
\end{table}

\section{Conclusion}
This paper introduces adversarial Voronoi patches, a novel form of adversarial
camouflage patterns that are not only indistinguishable from regular Voronoi
patterns to a human observer but also fool object detectors. Instead of
optimizing the patch itself, the proposed method optimizes positions of seed
points that are used to generate a Voronoi pattern using a fixed color palette.
The presented results strongly suggest that these patterns reduce a detector's
confidence significantly without the need of optimizing the pattern pixelwise.
Moreover, the transferability of the adversarial characteristic of the patterns
across multiple different architectures has also been shown.

The logical next step for the presented method is a realization and proof of
concept in the physical world. In addition, the not yet thoroughly examined
transferability across multiple datasets has also been taken into account.
Especially since all evaluated datasets are COCO-esque in terms of displaying
objects of interest considerably large and central in the image.

Future work should tackle these points and also consider extending the
investigations to different procedural methods for generating adversarial
camouflage patterns that resemble already used camouflage styles such as
Flecktarn or Fractals. Furthermore, there should be a more elegant way of
seamlessly lining up a patch instead of naive tiling. Hardening of the patterns
as well as investigating their impact on the activation maps of the detector
when present in the input data should be investigated as well.

\section*{Acknowledgments}
This work was developed in Fraunhofer Cluster of Excellence \enquote{Cognitive Internet Technologies}.

\bibliographystyle{abbrv}
\bibliography{references}  

@article{karras2021alias,
  title={Alias-free generative adversarial networks},
  author={Karras, Tero and Aittala, Miika and Laine, Samuli and H{\"a}rk{\"o}nen, Erik and Hellsten, Janne and Lehtinen, Jaakko and Aila, Timo},
  journal={Advances in neural information processing systems},
  volume={34},
  pages={852--863},
  year={2021}
}

@article{Peng2025,
author = {Peng, Zhenbang and Chen, Jianqi and Shi, Zhenwei and Zou, Zhengxia},
doi = {10.1109/TIFS.2025.3581771},
journal = {IEEE Transactions on Information Forensics and Security},
pages = {6308--6323},
publisher = {IEEE},
title = {{Physical Adversarial Camouflage Generation in Optical Remote Sensing Images}},
volume = {20},
year = {2025}
}

@article{Loshchilov2019,
archivePrefix = {arXiv},
arxivId = {1711.05101},
author = {Loshchilov, Ilya and Hutter, Frank},
eprint = {1711.05101},
journal = {7th International Conference on Learning Representations, ICLR 2019},
title = {{Decoupled weight decay regularization}},
year = {2019}
}

@article{Li2022,
  title={Bridging composite and real: towards end-to-end deep image matting},
  author={Li, Jizhizi and Zhang, Jing and Maybank, Stephen J and Tao, Dacheng},
  journal={IJCV},
  volume={130},
  number={2},
  pages={246--266},
  year={2022},
  publisher={Springer}
}

@inproceedings{Pumarola2019,
author = {Pumarola, Albert and Sanchez, Jordi and Choi, Gary and Sanfeliu, Alberto and Moreno-Noguer, Francesc},
booktitle = {ICCV},
title = {{3DPeople: Modeling the Geometry of Dressed Humans}},
year = {2019}
}

@inproceedings{Duan2020,
author = {Duan, Ranjie and Ma, Xingjun and Wang, Yisen and Bailey, James and Qin, A. K. and Yang, Yun},
booktitle = {CVPR},
doi = {10.1109/CVPR42600.2020.00108},
eprint = {1312.6034},
isbn = {978-1-7281-7168-5},
month = {jun},
pages = {997--1005},
publisher = {IEEE},
title = {{Adversarial Camouflage: Hiding Physical-World Attacks With Natural Styles}},
year = {2020}
}

@inproceedings{Duan2021,
author = {Duan, Yexin and Chen, Jialin and Zhou, Xingyu and Zou, Junhua and He, Zhengyun and Zhang, Jin and Zhang, Wu and Pan, Zhisong},
booktitle = {IJCAI},
doi = {10.24963/ijcai.2022/125},
eprint = {2109.00124},
isbn = {978-1-956792-00-3},
month = {jul},
pages = {891--897},
publisher = {International Joint Conferences on Artificial Intelligence Organization},
title = {{Learning Coated Adversarial Camouflages for Object Detectors}},
year = {2022}
}

@inproceedings{Zhu2022,
author = {Zhu, Xiaopei and Hu, Zhanhao and Huang, Siyuan and Li, Jianmin and Hu, Xiaolin},
booktitle = {CVPR},
doi = {10.1109/CVPR52688.2022.01296},
eprint = {2205.05909},
isbn = {978-1-6654-6946-3},
month = {jun},
pages = {13307--13316},
publisher = {IEEE},
title = {{Infrared Invisible Clothing: Hiding from Infrared Detectors at Multiple Angles in Real World}},
year = {2022}
}

@article{Wu2020,
author = {Wu, Zuxuan and Lim, Ser Nam and Davis, Larry S. and Goldstein, Tom},
doi = {10.1007/978-3-030-58548-8_1},
eprint = {1910.14667},
isbn = {9783030585471},
issn = {16113349},
journal = {Lect. Notes Comput. Sci.},
pages = {1--17},
title = {{Making an Invisibility Cloak: Real World Adversarial Attacks on Object Detectors}},
volume = {12349 LNCS},
year = {2020}
}

@article{Xu2020,
author = {Xu, Kaidi and Zhang, Gaoyuan and Liu, Sijia and Fan, Quanfu and Sun, Mengshu and Chen, Hongge and Chen, Pin Yu and Wang, Yanzhi and Lin, Xue},
doi = {10.1007/978-3-030-58558-7_39},
eprint = {1910.11099},
isbn = {9783030585570},
issn = {16113349},
journal = {LNCS)},
pages = {665--681},
title = {{Adversarial T-Shirt! Evading Person Detectors in a Physical World}},
volume = {12350 LNCS},
year = {2020}
}

@misc{Camogen2018,
    title= {camogen},
    note = {\url{http://www.happyponyland.net/camogen.php}, Accessed: 2023-09-01}
}

@inproceedings{Dalal2005,
author = {Dalal, N. and Triggs, B.},
booktitle = {CVPR},
doi = {10.1109/CVPR.2005.177},
isbn = {0-7695-2372-2},
issn = {2313433X},
pages = {886--893},
publisher = {IEEE},
title = {{Histograms of Oriented Gradients for Human Detection}},
volume = {1},
year = {2005}
}

@article{Lin2014,
archivePrefix = {arXiv},
arxivId = {1405.0312},
author = {Lin, Tsung Yi and Maire, Michael and Belongie, Serge and Hays, James and Perona, Pietro and Ramanan, Deva and Doll{\'{a}}r, Piotr and Zitnick, C. Lawrence},
doi = {10.1007/978-3-319-10602-1_48},
eprint = {1405.0312},
issn = {16113349},
journal = {LNCS},
number = {PART 5},
pages = {740--755},
title = {{Microsoft COCO: Common objects in context}},
volume = {8693 LNCS},
year = {2014}
}

@article{VanderBurg2025,
author = {{Van der Burg}, Erik and Toet, Alexander and Perone, Paola and Hogervorst, Maarten A.},
doi = {10.3390/app15095066},
issn = {20763417},
journal = {Appl. Sci.},
number = {9},
title = {{A Convolutional Neural Network as a Potential Tool for Camouflage Assessment}},
volume = {15},
year = {2025}
}

@article{Wei2024,
author = {Wei, Hui and Tang, Hao and Jia, Xuemei and Wang, Zhixiang and Yu, Hanxun and Li, Zhubo and Satoh, Shinichi and {Van Gool}, Luc and Wang, Zheng},
doi = {10.1109/TPAMI.2024.3430860},
eprint = {2209.15179},
issn = {19393539},
journal = {IEEE Trans. Pattern Anal. Mach. Intell.},
number = {12},
pages = {9797--9817},
publisher = {IEEE},
title = {{Physical Adversarial Attack Meets Computer Vision: A Decade Survey}},
volume = {46},
year = {2024}
}

@article{Chakraborty2021,
author = {Chakraborty, Anirban and Alam, Manaar and Dey, Vishal and Chattopadhyay, Anupam and Mukhopadhyay, Debdeep},
doi = {10.1049/cit2.12028},
issn = {24682322},
journal = {CAAI Trans. Intell. Technol.},
number = {1},
pages = {25--45},
title = {{A survey on adversarial attacks and defences}},
volume = {6},
year = {2021}
}

@article{Kim2023a,
author = {Kim, Jeonghun and Yang, Hunmin and Oh, Se-Yoon},
doi = {10.9766/KIMST.2023.26.1.044},
issn = {2636-0640},
journal = {KIMST},
month = {feb},
number = {1},
pages = {44--53},
title = {{Camouflaged Adversarial Patch Attack on Object Detector}},
volume = {26},
year = {2023}
}

@article{Wang2023a,
author = {Wang, Yang and Fang, Zheng and fei Zheng, Yun and Yang, Zhen and Tong, Wen and yong Cao, Tie},
doi = {10.1016/j.dt.2021.12.003},
issn = {22149147},
journal = {Defence Technology},
pages = {166--175},
title = {{Dual Attribute Adversarial Camouflage toward camouflaged object detection}},
volume = {22},
year = {2023}
}

@inproceedings{Bayer2024,
author = {Bayer, Jens and Becker, Stefan and M{\"{u}}nch, David and Arens, Michael},
booktitle = {Artif. Intell. Secur. Def. Appl. II},
doi = {10.1117/12.3031501},
editor = {Bouma, Henri and Yitzhaky, Yitzhak and Prabhu, Radhakrishna and Kuijf, Hugo J.},
eprint = {2408.15833},
isbn = {9781510681200},
month = {nov},
pages = {33},
publisher = {SPIE},
title = {{Network transferability of adversarial patches in real-time object detection}},
year = {2024}
}

\end{document}